\documentclass{article}
\usepackage{amssymb}
\usepackage{array}
\usepackage{booktabs}
\usepackage{amsmath}
\usepackage{floatrow}
\usepackage{adjustbox}
\usepackage{graphicx} 
\usepackage{enumitem}
\usepackage{color, soul} 

\usepackage[toc,page,header]{appendix}
\usepackage{minitoc}

\setlist[enumerate]{noitemsep, topsep=0pt}

\usepackage{xcolor}

\usepackage[final]{corl_2025} 

\title{Force-Modulated Visual Policy for Robot-Assisted Dressing with Arm Motions}

\author{
  Alexis Yihong Hao$^{1}$, \ Yufei Wang$^{1}$, \ Navin Sriram Ravie$^{2}$, \ Bharath Hegde$^{1}$, \\[3px]
  \textbf{David Held$^{\dagger1}$, \ Zackory Erickson$^{\dagger1}$} \\[3px]
  $^1${Robotics Institute, Carnegie Mellon University} \\
  $^2${Department of Engineering Design, Indian Institute of Technology, Madras} \\[3px]
  \small{\textbf{$\dagger$} Equal Advising}
}
\begin{document}
\maketitle
\vspace{-20px}


\begin{abstract}
    Robot-assisted dressing has the potential to significantly improve the lives of individuals with mobility impairments. To ensure an effective and comfortable dressing experience, the robot must be able to handle challenging deformable garments, apply appropriate forces, and adapt to limb movements throughout the dressing process. Prior work often makes simplifying assumptions---such as static human limbs during dressing---which limits real-world applicability. 
    In this work, we develop a robot-assisted dressing system capable of handling partial observations with visual occlusions, as well as robustly adapting to arm motions during the dressing process. 
    Given a policy trained in simulation with partial observations, we propose a method to fine-tune it in the real world using a small amount of data and multi-modal feedback from vision and force sensing, to further improve the policy's adaptability to arm motions and enhance safety. 
    We evaluate our method in simulation with simplified articulated human meshes and in a real world human study with 12 participants across 264 dressing trials. 
    Our policy successfully dresses two long-sleeve everyday garments onto the participants while being adaptive to various kinds of arm motions, and greatly outperforms prior baselines in terms of task completion and user feedback. Video are available at \url{https://dressing-motion.github.io/}.
    
\end{abstract}


\begin{figure}[h!] 
    \centering
    \includegraphics[width=.92\linewidth]{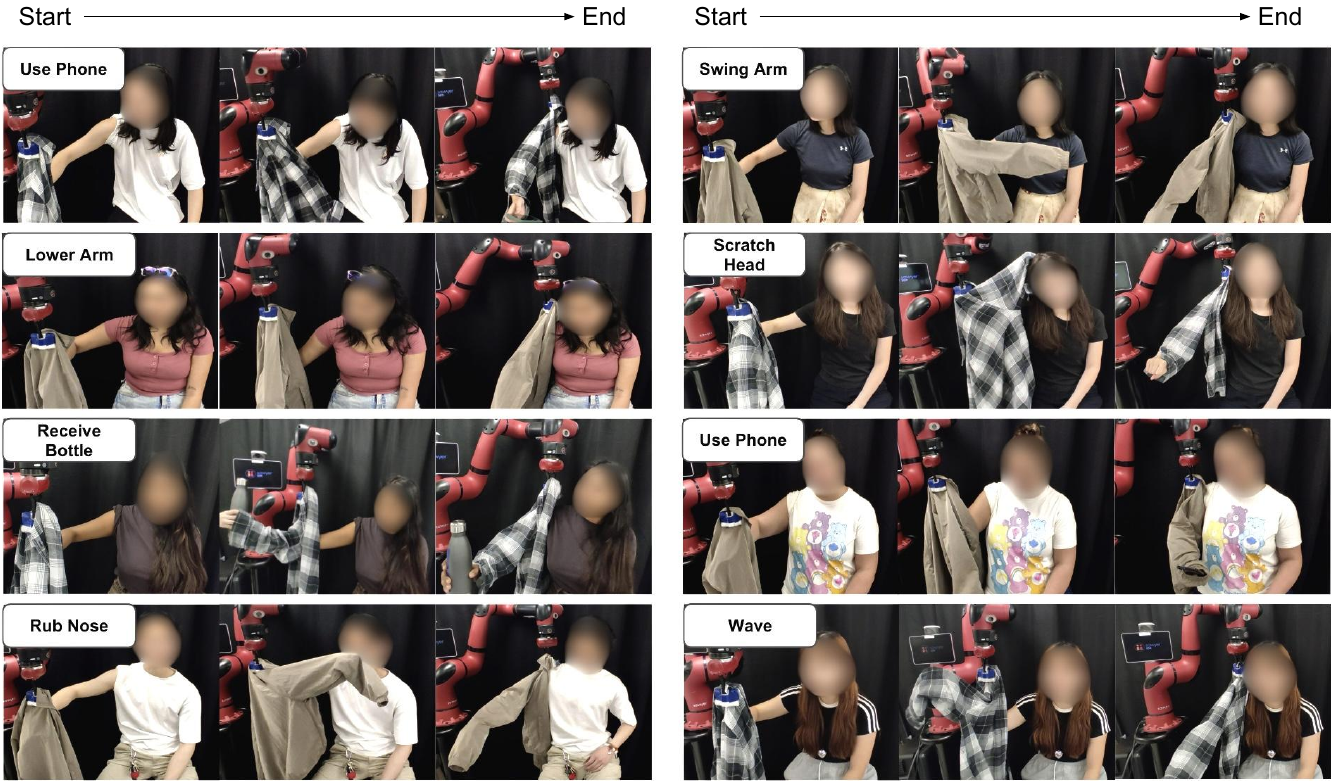}
    \caption{
     Snapshots from trajectories of our learned policy. It generalizes to dress different people with two everyday garments, while being robust to diverse arm motions during the dressing process. 
    }
    \label{fig:user-study-grid}
\end{figure}

\section{Introduction}
Although dressing is a fundamental daily activity, it remains a significant challenge for individuals with mobility impairments. A 2016 study by the National Center for Health Statistics~\citep{NCHS2019} reports that 92\% of nursing facility residents and at-home care recipients require caregiver assistance for dressing. Robot-assisted dressing systems have the potential to improve the quality of life and foster a greater sense of independence for these individuals. Such systems can also reduce caregiver workload, enabling caregivers to focus on tasks that still require human intervention.

However, robot-assisted dressing presents a multifaceted challenge. First, manipulating deformable garments is difficult due to the complex, non-linear dynamics of cloth and the absence of a compact state representation. Additionally, the gripper must operate in proximity to the person, applying force to the body through direct contact or indirectly via the garment. Undesired robot motions can cause discomfort by exerting excessive force or by causing the garment to get caught on the body. Finally, individuals with mobility impairments often struggle to hold their arm in a stable position that is convenient for dressing for extended time periods; therefore, a robust dressing system must adapt to user movement throughout the dressing process.

Several prior studies have investigated the problem of robot-assisted dressing in various settings. However, many approaches rely on assumptions that limit the generalizability of their systems. 
A common assumption in prior work is that the human limb remains static during the dressing process~\cite{erickson2018deep, qie2022cross, Wang2023One, sun2024force, kotsovolis2024garment}. 
While this simplifies the problem, it does not reflect the dynamic nature of human behavior.
Some approaches account for collaborative human motions that assist dressing~\citep{kapusta2019personalized, Ildefonso2021exploiting, clegg2020learning},  but they have not demonstrated robustness to arbitrary or disruptive limb movements—such as scratching an itch—that may interfere with the process.
In this paper, we aim to develop a robot-assisted dressing system that is robust to a broad spectrum of arm motions, including those that are non-cooperative, while also generalizing to two long-sleeve everyday garments and different people. 

The key to our approach for developing such a system is fine-tuning a simulation-trained policy using both visual and force feedback with a small amount of real-world data. 
We first train a vision-based policy in simulation using large-scale data under partial observations, enabling generalization across diverse body shapes and garment types.
However, this simulation training does not use any force information, as current simulators lack realistic force modeling for deformable garments.
Additionally, simulation training is conducted with static arms, as existing simulators are not yet stable or accurate enough to simulate deformable garment interactions with moving human limbs.
These limitations lead to a significant sim-to-real gap when deploying the policy directly on a real robot.
To address this, we propose Force-Modulated Visual Policy (FMVP), a new method for fine-tuning the policy in the real world using both vision and force feedback.
In particular, we condition the vision policy on force signals during fine-tuning, enabling the system to better adapt to dynamic arm motions while ensuring user safety.
We evaluate our method in a human study involving 12 participants across 264 dressing trials with varying garments and arm motions. On average, our method successfully dresses 85\% of each participant’s arm length.

In summary, the contributions of this paper are as follows:
\begin{itemize}
  \item We develop a new robot-assisted dressing system capable of adapting to non-cooperative arm motions and handling realistic garments with long sleeves.
  \item We propose a method that fine-tunes a simulation-trained policy using both visual and force feedback in the real world, improving adaptability to human motions.
  \item We conduct comprehensive evaluations in simulation and in a real-world human study with 12 participants and 264 dressing trials, demonstrating the effectiveness of our method across different garments and arm motions.
\end{itemize}


\section{Related Work}
\label{sec:relatedwork}
\subsection{Robot-Assisted Dressing}
Robot-assisted dressing has gained increasing attention in recent years. Early approaches often assume that the human remains static during dressing~\citep{erickson2018deep, qie2022cross, Wang2023One, sun2024force}, limiting adaptability to natural arm movements.
Collaborative frameworks~\citep{kapusta2019personalized, Ildefonso2021exploiting, gao2016iterative} instead rely on active user participation to facilitate dressing, but require sustained effort and are not designed for non-cooperative arm motions. Vision-only approaches~\citep{Wang2023One} ignore force sensing, risking excessive pressure in contact-rich interactions. A related method~\citep{sun2024force} augments a simulation-trained visual policy with real-world force sensing, but focuses on force minimization by predicting dynamics and filtering high-force actions, which may not always align with task success. We address these limitations by fine-tuning a pre-trained visual policy with real-world force feedback, enabling robust dressing under natural arm motions.
  
\subsection{Multi-Modal Learning for Robotic Manipulation}
Recent advances in multi-modal learning have improved robotic manipulation in complex, real-world settings by combining vision for spatial reasoning with force and tactile sensing for contact feedback.
These complementary modalities have been applied to tasks ranging from grasping, packing, pouring, and assistive tasks like dressing \citep{Yuan2023RobotSI, Watkins-Valls2019, sun2024force, Sunil2023, hu2024robocap, Wi2022, Lee2019}.
Early work in multi-modal learning focused on rigid object manipulation, where vision-touch fusion improved grasp stability, while more recent efforts extend to deformable objects by leveraging self-supervised and imitation learning to align visual, tactile, and force feedback during manipulation~\citep{Sundaresan2023, Li2023, Du2022}. However, simulating force and tactile signals for deformable objects remains challenging. Some methods address this by combining simulation-trained visual policies with real-world force-based dynamics models~\citep{sun2024force, gao2016iterative}. However, these methods treat vision and force as separate streams, using force primarily to predict unsafe actions or constrain motion, rather than learning a unified policy across modalities. In contrast, our method fine-tunes a visual policy by conditioning it directly on real-world force signals, enabling unified multi-modal policy learning without relying on separate dynamics models.

\section{Problem Statement and Assumptions}
\label{sec:problem}
As shown in Figure~\ref{fig:user-study-grid}, we study the task of single arm dressing with arm movements. The objective is to fully dress the garment's sleeve onto the person's arm, and the task is considered complete when the shoulder line of the garment is aligned with the participant's shoulder. 
Unlike prior work~\citep{sun2024force, Wang2023One} that assumes the person maintains a static arm pose during dressing, we remove this constraint. The goal of this paper is to develop a method that can robustly dress upper body garments despite arm movements during the dressing trial. We assume that the robot has already grasped the opening of the garment's shoulder in preparation for dressing, as grasping is not the focus of this paper. Prior work~\citep{jing2023grasp, zhang2020grasp2, zhang2022sciencerobotics} has introduced garment grasping techniques that could complement our method.

\section{Background - Vision-Based Policy Training in Simulation}
\label{sec:background}
The training of our vision-based policy in simulation is based on prior work~\citet{Wang2023One}, which we briefly review here. 
The policy is trained in NVIDIA FleX~\citep{agarwal2021flex} wrapped in SoftGym~\citep{lin2021softgym} using reinforcement learning (RL), formulating the dressing task as a Partially Observable Markov Decision Process (POMDP). Our policy directly takes the partial observation as input without explicitly estimating belief states, as commonly done for learning vision-based RL policies~\citep{dexpoint,ijcai2024p762}. Due to challenges in stably modeling cloth dynamics during interaction with a moving human arm in FleX, the training environment includes a range of static arm poses but no arm motion. The policy architecture is based on a segmentation-type PointNet++~\citep{qi2017pointnet++}, with SAC~\citep{haarnoja2018sac} being the RL algorithm.
The design of the POMDP is as follows: 

\textbf{Observation Space} \textit{O}: Each observation consists of a segmented point cloud representing the dressing scene, which includes the garment point cloud \textit{$P^g$}, the human arm point cloud \textit{$P^h$}, and a single point \textit{$P^r$} that represents the robot end-effector position. 
The full observation \textit{O} is the concatenation of the three types of points [\textit{$P^g$}, \textit{$P^h$}, \textit{$P^r$}], with each point annotated by a  one-hot feature indicating whether it belongs to the garment, human arm, or robot end-effector.
To get the segmented point cloud in the real world, \citet{Wang2023One} used color thresholding to segment the garment.
Since the arm was assumed to remain static, a complete arm point cloud could be captured before the dressing starts and used during the whole dressing process even when the arm became partially occluded by the garment. In contrast, our method uses a different approach for garment segmentation and accommodates dynamic arm movement during dressing. Both components are described below.

\textbf{Action Space} \textit{A}: The action is a 6D vector that represents the delta transformation of the robot end-effector, comprising three elements for delta translation and three for delta rotation in axis angle.

\textbf{Reward} \textit{r}: The reward includes multiple terms: a major term that measures task progress, which is quantified as the distance the garment has been dressed onto the arm, and several auxiliary terms to discourage the robot from moving too close to the person or exerting excessive force. See \citet{Wang2023One} for further details and a full formulation.

\section{Method}
\label{sec:method}

\begin{figure}[h!] 
    \centering
    \includegraphics[width=.9\linewidth]{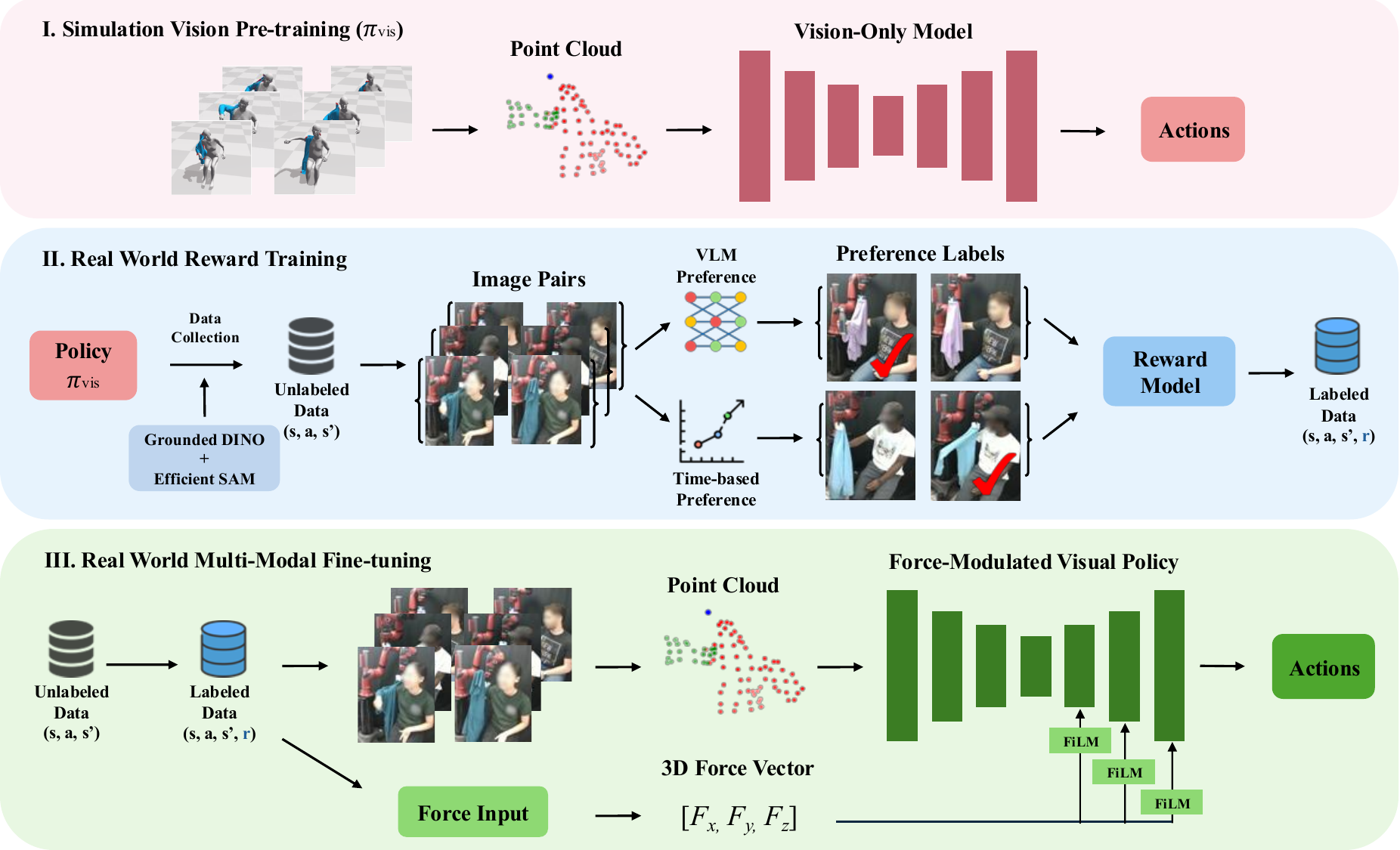}
    \caption{Overview of our method. (Top) We train a vision-based policy in simulation using reinforcement learning on a diverse range of human arm poses, garments, and body sizes. (Middle) We collect an unlabeled real-world dataset by rolling out the pre-trained policy, generate preference labels using a combination of VLM and time-based signals, and train a reward model to label the dataset. (Bottom) We fine-tune the simulation-pre-trained vision policy on a labeled real-world dataset using both vision and force information. Force signals are injected into the visual network via FiLM layers, which modulate the latent visual features.}
    \label{figure:method}
\end{figure}

Our method has three stages. 
First, we train a vision-based policy in simulation using reinforcement learning on a diverse range of human arm poses (albeit with static arms), garments, and body sizes following prior work~\cite{Wang2023One}. 
To adapt to arm motions, we remove the assumption in prior work that the complete arm can be observed despite garment occlusions during training. 
However, because the simulation policy is trained with occluded visual observations and without dynamic arm motions, it does not transfer well to real-world scenarios involving active limb movement.
To address this gap, we deploy the simulation-trained vision policy in a real-world human study to collect a small set of dressing trajectories with natural arm motions.
Finally,  we fine-tune the vision-based policy using the collected real-world data via offline RL with both vision and force feedback. 
When fine-tuning the policy, we condition the latent visual features on force inputs, enabling the policy to better adapt to arm movements during real-world dressing. 
Figure~\ref{figure:method} provides an overview of our method. 

\subsection{Vision-Based pre-training in Simulation with Partially Observable Point Cloud}

Since the goal of our method is to adapt to arm motion during dressing, we cannot rely on a pre-captured, complete arm point cloud as in other works~\citep{Wang2023One, sun2024force}. 
Instead, our observation consists of the garment point cloud \textit{$P^g$}, the visible (i.e., unoccluded) portion of the human arm point cloud \textit{$P^h_{vis}$}, and a single point to represent the robot end-effector position \textit{$P^r$}. In simulation, garment and arm point clouds can be  segmented using privileged information. 
We train a policy with this new observation space following the same reinforcement learning process as in~\citet{Wang2023One}. 

When transferring this policy to the real world, we segment the dressing garment using Grounding DINO~\citep{liu2023grounding} and EfficientSAM~\citep{xiong2023efficientsam}, and remove robot points from the scene with a fine-tuned Detectron2~\citep{wu2019detectron2} model.
Illustrations of the segmentation and masking process are provided in Appendix~\ref{app:pc_segmask}. 
To improve policy robustness, we distill a policy $\pi_{vis}$ using a filtered set of high-quality simulated trajectories. Please refer to Appendix~\ref{app:distill} for details on filtering and distillation.

\subsection{Multi-modal Fine-tuning in the Real World}
The simulation trained vision policy $\pi_{vis}$ may not transfer zero-shot to real world settings with arm motions for two main reasons.
1) During dressing, only the unoccluded portion of the arm is visible, making it difficult to infer the arm’s position beneath the garment from point clouds alone. Subtle changes in elbow angle, for example, can cause the garment to snag, yet these movements are often unobservable by the camera. This challenge is exacerbated by noisy real-world point clouds caused by e.g., segmentation errors. 
Force feedback offers additional information about the interactions between the garment and the arm, mitigating the limitations of vision alone. However, current simulators do not offer accurate enough force modeling for deformable objects in contact with human limbs.
2) In simulation, the policy is trained on static human meshes, since cloth simulation with actuated humans in NVIDIA FleX is unstable and limited in fidelity. Consequently, real-world arm motions create out-of-distribution states that the policy has never encountered, resulting in failures during execution.
To address these challenges, we collect a small amount of real-world data with natural arm movements and fine-tune the simulation-trained policy. This fine-tuning incorporates both visual and force feedback, improving robustness to dynamic arm motions.

Specifically, we rollout policy $\pi_{vis}$ to collect a set of sub-optimal real-world dressing trajectories with arm movements $D = \{\tau_i\}_{i=1}^N$. Each trajectory consists of $T$ observation-action-reward pairs $\tau = \{o_i, a_i, r_i\}_{i=1}^T$. The observation includes both the segmented point cloud observation as well as a force measurement $f \in \mathbb{R}^{3}$.
The reward contains two terms, where the first measures the task progress, i.e., how much the garment has been dressed onto the human arm, and the second ensures safety that penalizes excessive force being applied to the human body. We describe in more detail how this dataset is collected in the real world in Section~\ref{subsec:method-reward}. 
We use Implicit Q Learning (IQL)~\cite{kostrikov2021iql} as the underlying offline RL algorithm to perform fine-tuning of $\pi_{vis}$ with this offline dataset $D$. 

We now describe how we incorporate the force information into the vision policy. 
Specifically, the force vector $f \in \mathbb{R}^{3}$ is passed to a set of FiLM layers~\citep{perez2018film}, one for each feature propagation layer of the PointNet++ network used in the vision policy. 
Each FiLM layer $i$ produces two modulation vectors: $\gamma_i \in \mathbb{R}^{d_i}$ and $\beta_i \in \mathbb{R}^{d_i}$, where $d_i$ denotes the feature dimension at feature propagation layer $i$ in the PointNet++ network. 
These vectors are then broadcasted to match the number of points in the feature propagation layer, resulting in $\Gamma_i \in \mathbb{R}^{n \cdot d_i}$ and $B_i \in \mathbb{R}^{n \cdot d_i}$, where $n$ is the number of points. 
Following the approach proposed by Shridhar et al.~\citep{shridhar2022film}, we apply FiLM conditioning to every feature propagation layer of the PointNet++ network: the feature map $F_i$ at layer $i$ is modulated using the conditioning such that $F'_i$ = $\Gamma_i \odot F_i$ + $B_i$, where $\odot$ represents a Hadamard product.
The force measurements at the robot end-effector can be noisy as they are derived from joint torques. To address this, we apply exponential moving average of the force vector to smooth it. 

\subsection{Data collection and Reward Labeling in the Real World}
\label{subsec:method-reward}
To collect the real world dataset $D$ for fine-tuning the simulation-trained vision policy $\pi_{vis}$, 
we run a user study with 8 participants, including 5 males and 3 females, with ages from 21 to 29. For each participant, we collect data across 3 garments and 8 arm motions by running the policy $\pi_{vis}$, resulting in 24 trials per participant and 192 trials in total (See Fig.~\ref{fig:human-study-setup} for the garments and motions). We record both segmented point clouds and force measurements when collecting data. 

During simulation training, we compute the reward term that measures dressing progress using ground truth information in the simulator (e.g., garment particle positions). 
However, such information is not directly accessible in the real world. To obtain real-world rewards for dressing progress from image observations, we adopt RL-VLM-F~\citep{wang2024rlvlmfreinforcementlearningvision, venkataraman2024offlinerlvlmf}, which queries a vision-language model (VLM) for preference labels between randomly sampled image pairs based on how well they achieve the task goal of “successfully dressing the jacket onto the arm.”
The VLM outputs a preference label $l \in \{-1, 0, 1\}$ indicating which image better achieves the goal.
To supplement this, we also introduce time-based preference labels: given an image pair ($\text{I}_i$, $\text{I}_j$) from time step $i$ and $j$ within a trajectory, the model assigns a preference label of 0 if $i > j$, 1 if $i < j$ and -1 if $i = j$. This assumes steady progress during the trial, so images from later time steps are generally preferred. 
While time-based labels are effective for successful trajectories, they are unreliable for failed trials where the garment is caught on the arm or where the robot performs undesirable actions that hinder dressing. Conversely, VLM-generated labels can be noisy and inconsistent in ambiguous scenarios. 
To balance their strength, we combine 4000 VLM-generated labels with 4000 time-based labels to train a reward model using the Bradley-Terry formulation~\citep{bradleyterry1952}, which is then used to label the entire real-world dataset $D$. In addition to this task progress reward, we include a force penalty term to penalize excessive applied forces. Details on reward model training are provided in Appendix~\ref{app:reward_training}.

\section{Simulation Experiments}
\label{sec:simexp}

\subsection{Sim-to-Sim Transfer Setup}
\label{subsec:sim-setup}
We first evaluate our method in a sim-to-sim transfer setting by creating a second simulation environment  using Assistive Gym~\citep{erickson2020assistivegym}, built on the PyBullet simulator~\citep{coumans2021pybullet}, which supports simplified actuated human meshes with cylindrical limb. Although PyBullet still lacks accurate simulation for deformable cloth interacting with actuated human arms, it provides a controlled setting for testing methods. The force readings and garment dynamics in Assistive Gym differ from FleX where the vision-based policy $\pi_{vis}$ is trained, approximating a sim-to-real gap. 
In PyBullet, we generate four different body sizes---small, medium, large, and extra large---by varying arm length and arm radius.

We define 14 arm motions by executing seven distinct arm motions and replaying each in reverse (see Table~\ref{table:sim-motion}), and select three garments from the Cloth3D dataset~\citep{Bertiche2020cloth3d} with different sleeve widths and lengths. We collect a dataset of 204 trajectories using $\pi_{vis}$ on the medium body size, a subset of five arm motions, and two garments in PyBullet. We then fine-tune $\pi_{vis}$ using this dataset and perform evaluations on all body sizes, arm motions, and garments in PyBullet. 

For each trial during data collection and evaluation, the initial arm configuration is randomized by adding offsets of up to ±10 cm per axis at the elbow and up to ±15 cm per axis at the hand from their default positions in Assistive Gym. For each combination of method, garment, arm motion, and body size, we run ten randomized trials and report the average, resulting in 3 $\times$ 14 $\times$ 4 $\times$ 10 = 1680 evaluation trials per method. Details on the simulation experiment setup are provided in Appendix~\ref{app:sim_setup}. 
Following prior work~\citep{Wang2023One, sun2024force}, we use the \textbf{Upper Arm Dressed Ratio} as the evaluation metric, defined as the ratio between the dressed upper arm length to the true upper arm length.

\subsection{Baselines and Ablations}
\label{sec:sim-baseline}
We compare the following methods and ablations, which differ in their use of visual and force feedback for policy learning and adaptation. 
\textbf{FMVP (Ours)} is our proposed method, described in Section~\ref{sec:method}. 
\textbf{Vision-based Policy $\pi_{vis}$} is trained in NVIDIA FleX using only visual observations and transferred to PyBullet without adaptations. 
\textbf{FCVP~\citep{sun2024force}} uses $\pi_{vis}$ to propose actions and trains a force dynamic model in PyBullet to filter out actions that would exceed a predefined force threshold. 
\textbf{Scratch-IQL (FiLM)} is trained from scratch using the dataset of 204 trials collected in PyBullet. It uses the same PointNet++ architecture and FiLM layers to incorporate force information as in our method. 
\textbf{Scratch-IQL (Concat)} is also trained from scratch using the dataset of 204 trials collected in PyBullet. It uses the same PointNet++ architecture as our method, but instead of FiLM conditioning, it concatenates the force magnitude to the robot end-effector position as an additional input feature.
\textbf{Vision Fine-tuning} only fine-tunes the vision network of $\pi_{vis}$ in PyBullet and has no FiLM layers. 
\textbf{Force Fine-tuning} follows the same approach as our method, except the vision encoder of $\pi_{vis}$ is frozen during fine-tuning. 
\textbf{BC Fine-tuning} follows the same approach as our method, except Behavioral Cloning (BC)~\citep{torabiBC} is used as the underlying fine-tuning algorithm.

\subsection{Simulation Results}
\label{subsec:sim-results}

\begin{table}[h]
\scriptsize
\centering
\begin{tabular}[t]{l
    >{\centering\arraybackslash}m{0.1\linewidth}
    >{\centering\arraybackslash}m{0.1\linewidth}
    >{\centering\arraybackslash}m{0.1\linewidth}
    >{\centering\arraybackslash}m{0.1\linewidth}
    | >{\centering\arraybackslash}m{0.1\linewidth}
}
\toprule
 & Small & Medium & Large & Extra Large & Average\\
\midrule
\textbf{FMVP (Ours)}                      & \textbf{0.63}   & \textbf{0.71}   & \textbf{0.62}   & \textbf{0.61}   & \textbf{0.64} \\
Vision-based              & 0.42   & 0.42   & 0.29   & 0.32   & 0.36 \\
FCVP                      & 0.42   & 0.43   & 0.29   & 0.32   & 0.37 \\
Scratch-IQL (Film)        & 0.54   & 0.51   & 0.51   & 0.39   & 0.49 \\
Scratch-IQL (Concat)      & 0.53   & 0.56   & 0.40   & 0.41   & 0.47 \\
Vision-only Fine-tuning   & 0.50   & 0.49   & 0.36   & 0.41   & 0.44 \\
Force-only Fine-tuning    & 0.55   & 0.60   & 0.44   & 0.37   & 0.49 \\
BC Fine-tuning            & 0.56   & 0.54   & 0.45   & 0.26   & 0.45 \\

\bottomrule
\end{tabular}

\caption{Upper arm dressed ratio of all methods across different body sizes.
}
\label{table:sim-body-size}
\end{table}

\begin{table}[ht]
\centering
\begin{adjustbox}{max width=\textwidth}
\begin{tabular}{
    l
    *{14}{>{\centering\arraybackslash}c}
}
\toprule
& \shortstack{raise\\arm} 
& \shortstack{lower\\arm} 
& \shortstack{open\\arm} 
& \shortstack{reach\\pocket} 
& \shortstack{reach\\side} 
& \shortstack{scratch\\head} 
& \shortstack{reach\\up} 
& \shortstack{rev.\\raise\\arm} 
& \shortstack{rev.\\lower\\arm} 
& \shortstack{rev.\\open\\arm} 
& \shortstack{rev.\\reach\\pocket} 
& \shortstack{rev.\\reach\\side} 
& \shortstack{rev.\\scratch\\head} 
& \shortstack{rev.\\reach\\up} \\
\midrule
\textbf{FMVP (Ours)}                       & \textbf{0.77} & \textbf{0.70} & \textbf{0.78} & \textbf{0.82} & \textbf{0.80} & \textbf{0.35} & \textbf{0.83} & \textbf{0.62} & 0.63 & \textbf{0.68} & \textbf{0.33} & \textbf{0.84} & \textbf{0.44} & \textbf{0.43} \\
Vision-based               & 0.45 & 0.43 & 0.41 & 0.38 & 0.44 & 0.32 & 0.36 & 0.34 & 0.36 & 0.38 & 0.23 & 0.56 & 0.28 & 0.17 \\
FCVP                       & 0.44 & 0.44 & 0.41 & 0.37 & 0.45 & 0.33 & 0.38 & 0.33 & 0.36 & 0.38 & 0.23 & 0.59 & 0.26 & 0.16 \\
Scratch-IQL (FiLM)         & 0.57 & 0.60 & 0.62 & 0.70 & 0.71 & 0.26 & 0.60 & 0.45 & 0.68 & 0.53 & 0.23 & 0.66 & 0.02 & 0.17 \\
Scratch-IQL (Concat)       & 0.68 & 0.46 & 0.54 & 0.65 & 0.73 & 0.14 & 0.35 & 0.55 & \textbf{0.72} & 0.62 & 0.16 & 0.80 & 0.13 & 0.12 \\
Vision-only Fine-tuning    & 0.57 & 0.53 & 0.54 & 0.49 & 0.55 & 0.28 & 0.41 & 0.53 & 0.67 & 0.44 & 0.27 & 0.67 & 0.03 & 0.16 \\
Force-only Fine-tuning     & 0.67 & 0.65 & 0.65 & 0.74 & 0.69 & 0.25 & 0.48 & 0.45 & 0.63 & 0.53 & 0.23 & 0.73 & 0.01 & 0.19 \\
BC Fine-tuning     & 0.47 & 0.55 & 0.59 & 0.73 & 0.64 & 0.17 & 0.51 & 0.61 & 0.46 & 0.56 & 0.16 & 0.57 & 0.05 & 0.29 \\

\bottomrule
\end{tabular}
\end{adjustbox}
\caption{Upper arm dressed ratio of all methods across different arm motions. ``rev." denotes the reversed version of the original motion.} 
\label{table:sim-motion}
\end{table}

Table~\ref{table:sim-body-size} and ~\ref{table:sim-motion} report the performance of all methods and ablations. FMVP achieves the highest upper arm dressed ratio on 13 of 14 arm motions and across all body sizes, outperforming the baselines by 0.15-0.28. Notably, while all other methods show clear performance degradation as body size increases from Medium (the training body size) to Large and Extra Large, FMVP maintains consistent performance across all three unseen body sizes.

The large gap between FMVP and Vision-based Policy highlights the value of fine-tuning in the target environment. Comparison with FCVP further shows the benefit of incorporating force feedback while directly optimizing for the task objective. FCVP predicts next-step forces and filters high-force actions, but this is less effective under arm motion, where future arm positions are uncertain and the garment may be pulled in ways that induce unanticipated force. By conditioning on force feedback and directly optimizing for dressing success, FMVP achieves more robust performance. 

FMVP also outperforms Vision-only and Force-only Fine-tuning, indicating that leveraging both modalities during fine-tuning provides greater gains than using either alone. Additionally, the gap between FMVP and Scratch-IQL (FiLM) demonstrates the importance of pre-training in simulation across diverse arm poses, garments, and body sizes. This is evident as FMVP maintains similar performance across all three unseen body sizes, while the performance of Scratch-IQL (FiLM) drops substantially on Extra Large. Finally, among fine-tuning strategies, BC Fine-tuning performs worse than IQL, likely because the dataset collected by the pre-trained vision policy is suboptimal. In contrast, our RL approach can learn beyond the limitations of demonstration data.

\section{Real-World Experiments and Human Study}
\label{sec:realexp}
\subsection{Human Study Setup}
\begin{figure}[!htb] 
    \centering
    \includegraphics[width=\textwidth]{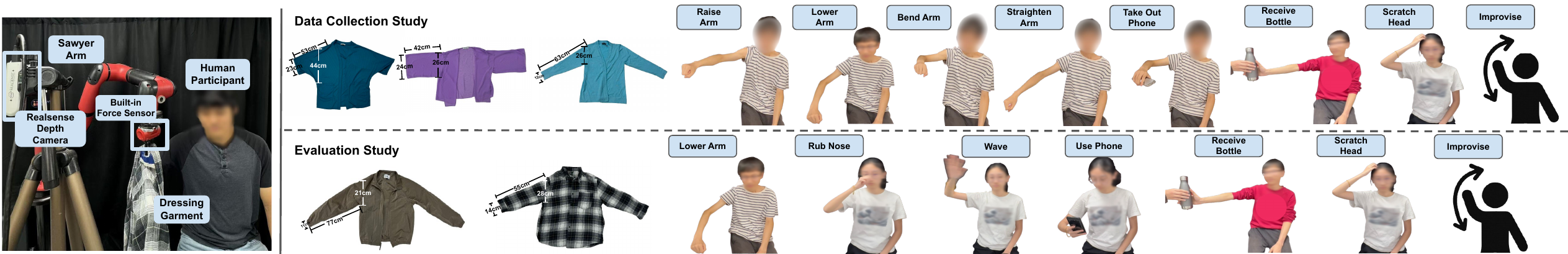}
    \caption{Human study setup (left), garments (middle) and arm motions (right) used in the studies. }
    \label{fig:human-study-setup}
\end{figure}

Figure~\ref{fig:human-study-setup} shows the setup of our real-world human study (left), dressing garments (middle), and arm motions (right). The motions are designed with large movements as stress tests to evaluate policy robustness to unpredictable arm behaviors such as tremors, spasticity, or posture shifts, which can lead to complex garment-body interactions. Four of the seven arm motions and both garments used in the evaluation study are not included in the data collection study for fine-tuning our method. Dressing is performed with a Sawyer arm, which provides force readings at the end-effector via built-in force sensors, and an Intel RealSense D435i camera is used to capture the point cloud. Impedance control is applied during the study to ensure safe interactions with participants.

We evaluate each dressing trial using the \textbf{Whole Arm Dressed Ratio}, defined as the ratio of the dressed arm length to the true whole arm length, and \textbf{Upper Arm Dressed Ratio} (defined in Section~\ref{subsec:sim-setup}). At the end of each trial, participants respond to four 7-point Likert items (1 = ``Strongly Disagree", 7 = ``Strongly Agree"): 1) ``The robot successfully dressed the garment onto my arm''; 2) ``The force the robot applied to me during dressing was appropriate''; 3) ``The dressing process was comfortable for me''; 4) ``The robot was robust to my arm motion during dressing''.
We compare FMVP against two baselines: Vision-based Policy and FCVP~\cite{sun2024force}.

\subsection{Human Study Procedure}
We recruit 12 participants (5 males and 7 females, age 21-39). For each participant, we conduct 11 trials per garment, for a total of 22 trials. Of the 11 trials per garment, our method is evaluated on all seven arm motions shown in Figure~\ref{fig:human-study-setup} (bottom), while each baseline is evaluated on two randomly selected motions from the set of seven. 
Based on the feedback from data collection study sessions, most participants experience arm fatigue towards the end of the study. Therefore, we find it impractical to evaluate both baselines on all arm motions and limit the number of trials to 22. The ordering of the methods, arm motions, and garments are counterbalanced for each participant.

For six of the seven arm motions, participants watch a demonstration video and mimic the motion; for the remaining motion, they perform an improvised arm motion without demonstration. This condition is included to evaluate the robustness of our method to unpredictable arm movements, which can occur during real-world dressing scenarios. Each trial stops if one of the following criteria is met: (1) the policy runs up to 80 steps, (2) the participant's shoulder is covered by the garment, (3) the robot stops making dressing progress for more than 10 consecutive steps, (4) the participant requests to stop, or (5) the force exceeds the safety threshold of 18N, taking reference from~\citep{sun2024force}.

\subsection{Human Study Results and Analysis}
\begin{figure}[h!]
\centering
\begin{minipage}[c]{0.6\linewidth}
    \centering
    \includegraphics[width=\linewidth]{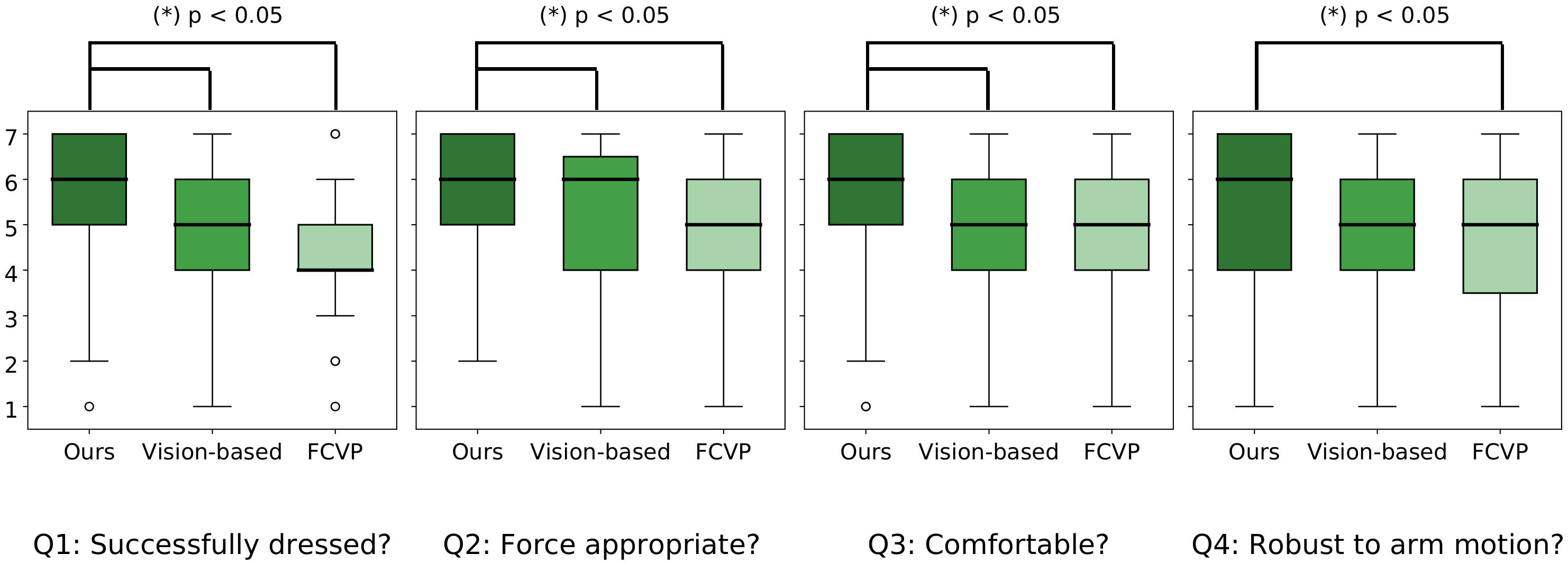}
\end{minipage}%
\scriptsize
\begin{minipage}[c]{0.4\linewidth}
    \centering
    \begin{tabular}{l 
    >{\centering\arraybackslash}m{0.2\linewidth}
    >{\centering\arraybackslash}m{0.25\linewidth}
    }
        \toprule
         & \shortstack{Upper Arm\\Dressed Ratio} & \shortstack{Whole Arm\\Dressed Ratio} \\
        \midrule
        \textbf{FMVP(Ours)}  & \textbf{0.79} & \textbf{0.86}\\
        Vision-based & 0.63 & 0.81\\
        FCVP & 0.50 & 0.73\\
        \bottomrule
    \end{tabular}
\end{minipage}
\caption{Likert item responses (left) and average arm dressed ratios (right), evaluated on the 48 trials where the same arm motions and garments are tested for all methods.}
\label{fig:human_study_compare}
\end{figure}
Figure~\ref{fig:user-study-grid} shows snapshots of dressing trials using FMVP in the human study. Videos are available on our project website. In Figure~\ref{fig:human_study_compare}, we compare FMVP against the two baselines on the 48 trials where all methods are evaluated on the same arm motions and garments. Our method achieves the highest arm dressed ratios, and the Likert item responses indicates that participants generally agree that FMVP provides a better dressing experience. 
On average, participants ``Agree" that FMVP successfully dressed the garment, applied appropriate force, ensured a comfortable experience, and was robust to arm motions. In contrast, the two baselines achieve median scores of 4.0 and 5.0, meaning that participants ``Somewhat Agree" or are ``Neutral" about these statements. Across the full set of 168 trials, FMVP attains an average whole arm and upper arm dressed ratio of 0.85 and 0.74, respectively. Further analysis is provided in Appendix~\ref{app:more-eval-analysis}.

\section{Conclusion}
\label{sec:conclu}
We present a robot-assisted dressing system that robustly adapts to diverse arm motions. By fine-tuning a simulation-trained vision policy with a small set of multi-modal real-world data, our approach improves adaptability and safety. Extensive evaluations in simulation and a real-world study show that our method outperforms prior baselines, enabling reliable and comfortable assistance.

\section{Limitation}
\label{sec:limit}
One limitation of our work is the assumption that the robot has already grasped the garment before each trial. Prior work for learning garment grasping~\citep{jing2023grasp, zhang2020grasp2, zhang2022sciencerobotics} can be combined with our system to relax this assumption. We also assume the participant's arm starts at a position accessible to the robot, and arm motions occur primarily after the arm is partially inserted into the sleeve. This helps simplify the experimental setup and reflects common scenarios where users position their arm to initiate dressing, then naturally move during the process. Such assumptions are also used in prior work~\citep{Wang2023One, kapusta2019personalized}. Our work can potentially be combined with methods for limb repositioning or initial limb alignment to address cases where the arm begins hanging down. Additionally, using only a single camera in the real-world setup often leads to occlusions and missing regions in the point cloud. This could be mitigated by incorporating multiple cameras or active sensing strategies, where the camera actively moves to capture views that minimize occlusion. Another limitation is the slow trial speed: each robot step takes about one second, resulting in trials lasting up to 80 seconds. The primary bottleneck is the inference time of Grounded DINO. Although we experimented with faster tracking and detection models~\citep{bekuzarov2023xmem, Cheng2024YOLOWorld}, they degraded performance by failing to capture shadows along clothing folds, introducing gaps into the point cloud. Finding faster yet accurate segmentation methods remains an important direction for future work. While the inference speed is low, it does not constrain participant's motion timing. During the user study, participants were instructed to perform natural, realistic arm motions, which introduced meaningful variations in garment dynamics, arm pose, and body-garment interactions. Finally, force readings at the robot end-effector, estimated from joint torques, may be noisy; this could be addressed by incorporating a dedicated force-torque sensor.

\clearpage
\acknowledgments{Research reported in this publication was supported in part by National Institute of Biomedical Imaging and Bioengineering of the National Institutes of Health under award number 1R01EB036842-01, the National Science Foundation under NSF CAREER Grant No. IIS-2046491, and NIST under Grant No. 70NANB24H314. Any opinions, findings, and conclusions or recommendations expressed in this material are those of the author(s) and do not necessarily reflect the views of National Institutes of Health, National Science Foundation, or NIST. We would like to thank Mino Nakura, Yishu Li, and Chenyuan Hu, Divyam Goel, and Sriram Krishna for helping with pilot studies. }


\bibliography{example}  

\doparttoc 
\faketableofcontents 
\appendix
\clearpage
\onecolumn
\addcontentsline{toc}{section}{Appendix} 
\part{Appendix} 

\parttoc 
\section{System Implementation Details}
\subsection{Simulation Policy Distillation}
\label{app:distill}
As mentioned in the main paper Section 5.1, we distill the simulation-trained vision-based policy using a filtered set of high-quality trajectories to improve robustness. This section provides details on the data collection, filtering criteria and distillation progress. 

We begin by rolling out the policy trained with partial observations using reinforcement learning (as detailed in Section 5.1 in the main paper) in the NVIDIA FleX~\citep{agarwal2021flex} simulation environment, following the setup used in prior work~\citep{Wang2023One}. 
The environment includes 27 arm pose regions, with 5 distinct arm poses per region, and 5 garments. For each trial, we randomly sample a region-pose-garment combination and collect the state-action pairs ($s_i$, $a_i$) at each time step $i$. In total, we collect more than 8000 trajectories. To ensure quality, we filter the trajectories using the following two criteria:
\begin{enumerate}
    \item The upper arm dressed ratio must be at least 0.7
    \item The trajectory must not exhibit \textit{early turning} behavior around the elbow
\end{enumerate}
We define an \textit{early turn} as any step where the gripper enters the inner side of the arm while in the elbow region. The elbow region is defined as the union of the back 1/4 segment of the forearm and the front 1/4 segment of the upper arm. To determine whether the gripper is on the inner side of the arm, we use 2D cross products in the XZ-plane. Specifically, we first define the following vectors:
\begin{itemize}
    \item $\vec{v}_1 = \texttt{hand\_pos} - \texttt{gripper\_pos}$: the vector from the gripper to the hand
    \item $\vec{d}_1 = \texttt{hand\_pos} - \texttt{elbow\_pos}$: the direction from the elbow to the hand
    \item $\vec{v}_2 = \texttt{elbow\_pos} - \texttt{gripper\_pos}$: the vector from the gripper to the elbow
    \item $\vec{d}_2 = \texttt{elbow\_pos} - \texttt{shoulder\_pos}$: the direction from the shoulder to the elbow
\end{itemize}

We then compute the signed scalar values of the following 2D cross products (in the XZ-plane):
\begin{itemize}
    \item $c_1 = (\vec{v}_1 \cdot \vec{d}_1)$
    \item $c_2 = (\vec{v}_2 \cdot \vec{d}_2)$
\end{itemize}

\begin{figure}[h!] 
    \centering
    \vspace{-15mm}
    \includegraphics[width=0.6\linewidth]{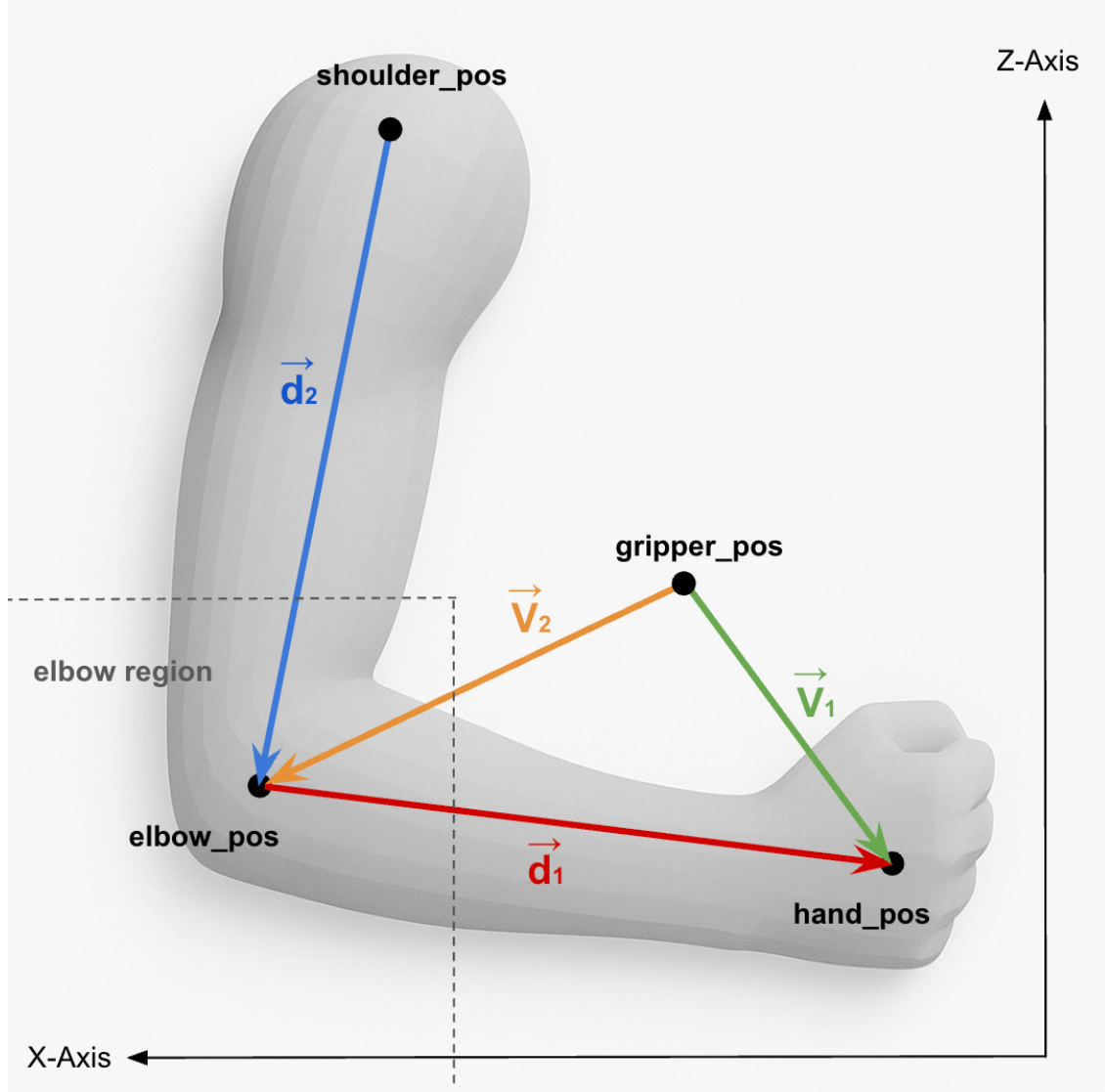}
    \vspace{-15mm}
    \caption{
     Key spatial information used to define an early turn.
    }
    \label{fig:early-turn}
    \vspace{-2mm}
\end{figure}

If both $c_1 < 0$ and $c_2 < 0$, the gripper is considered to be on the inner side of the arm. A trajectory is flagged as containing an early turn if the gripper is on the inner side for at least one time step within the elbow region.

After filtering, we obtain a total of 2514 high-quality trajectories, each with an upper arm dressed ratio of at least 0.7 and no early turning behavior. We use this filtered dataset to distill a policy via behavior cloning. The policy network follows that of~\citep{Wang2023One}, which is a segmentation-type PointNet++~\citep{qi2017pointnet++} consisting of:
\begin{itemize}
    \item Two set abstraction layers with radii of 0.05 and 0.1, and sampling ratio of 1.0 for both
    \item A global max pooling layer
    \item Three feature propagation layers, with 1, 3, and 3 nearest neighbors
    \item A multi-layer perception (MLP) for final action prediction
\end{itemize}

We train the policy using the Adam optimizer~\citep{Kingma2014AdamAM} with learning rate of $1 \times 10^{-4}$, and a batch size of 128. We train the policy by minimizing the negative log likelihood of the action on the high-quality trajectories. 
The final checkpoint used is trained for 40,000 steps. 

\subsection{Real-World Point Cloud Segmentation and Masking}
\label{app:pc_segmask}
We process the real-world point cloud using Grounding DINO~\citep{liu2023grounding} combined with EfficientSAM~\citep{xiong2023efficientsam} to segment the dressing garment, and Detectron2~\citep{wu2019detectron2} to segment and mask the Sawyer robot arm. For all five of our garments used in our data collection and evaluation studies, we use ``cloth" as the text prompt for Grounding DINO.

To segment the Sawyer arm, we manually label 50 images from earlier dressing trials with binary arm masks. Of these, 47 images are used for training and 3 for validation. We fine-tune the Mask R-CNN Restnet-50~\citep{matterport_maskrcnn_2017, he2016resnet} model from Detectron2, which is pre-trained on the COCO dataset~\citep{cocodataset}, for 400 epochs on the 47 training images.

\begin{figure}[h!] 
    \centering
    \vspace{-25mm}
    \includegraphics[width=0.6\linewidth]{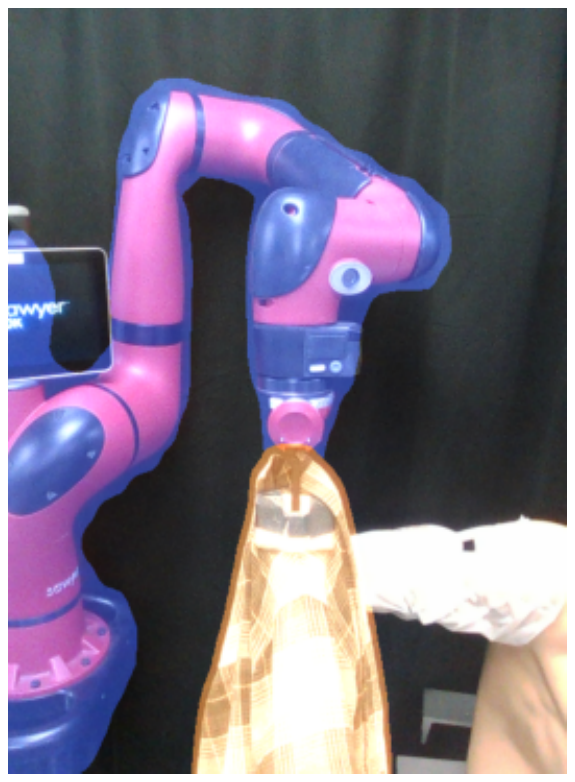}
    \vspace{-25mm}
    \caption{
     Robot (blue) and garment (orange) segmentation masks after dilation.
    }
    \label{fig:mask}
    \vspace{-2mm}
\end{figure}

To ensure full coverage at the boundaries of the garment and robot, we dilate their respective segmentation masks by 11 pixels. 
See Figure~\ref{fig:mask} for an illustration of the garment and robot masks.

\subsection{Real-World Preference Reward Comparison}
\label{app:pref_reward_eval}
As described in Section 5.3, we generate preference labels on real-world image pairs using a combination of vision-language model (VLM) labels and time-based labels. To study the effect of different labeling strategies, we conduct ablation experiments comparing three variants of the real-world reward model, each trained with a different labeling scheme:

\begin{itemize}
    \item \textbf{Ours:} 4000 VLM labels + 4000 time-based labels
    \item \textbf{Time-only:} 8000 time-based labels
    \item \textbf{VLM-only:} 8000 VLM labels
\end{itemize}

We fine-tune a policy using each of the reward models, following the procedure described in Section 5.2. We test all three policies with three participants, randomly selecting one garment and three arm motions per participant. For each participant, the same three motions are used across all three methods. The motion assignments are counterbalanced across participants to ensure that each arm motion is tested at least once with each method. As in the main evaluation study, we report the average whole arm dressed ratio and average upper arm dressed ratio as our evaluation metrics. The results are presented in Table~\ref{table:reward-model-eval}.
As shown, using a mixture of the VLM and time-based preference labels leads to the best performance. 

\begin{table}[h]
\centering
\begin{tabular}{l
    >{\centering\arraybackslash}m{0.2\linewidth}
    >{\centering\arraybackslash}m{0.2\linewidth}}
\toprule
 & Upper Arm Dressed Ratio & Whole Arm Dressed Ratio \\
\midrule
VLM+Time-based (Ours)     & \textbf{0.73}   & \textbf{0.87} \\
Time-based Only           & 0.61   & 0.77 \\
VLM Only                  & 0.66   & 0.84 \\
\bottomrule
\end{tabular}
\vspace{0.4em}
\caption{Arm dressed ratio of policies labeled using different reward models}
\label{table:reward-model-eval}
\end{table}

\subsection{Real-World Reward Model Training}
\label{app:reward_training}
We follow the preference-based reward learning framework described in~\citep{venkataraman2024offlinerlvlmf}. In our setting, a reward function is learned from preferences over agent behavior, where preference labels are automatically generated using either a VLM or a time-based heuristic. Formally, a segment $\sigma$ is defined as a sequence of states: $\{s_t\}_{i=1}^H$. In our case, we simplify each segment to a single image. Given a pair of segments $(\sigma_0, \sigma_1)$, an annotator provides a preference label $y \in \{-1, 0, 1\}$:
\begin{itemize}
    \item $y = 0$ indicates that $\sigma_0$ is preferred,
    \item $y = 1$ indicates that $\sigma_1$ is preferred,
    \item $y = -1$ indicates no preference (incomparable).
\end{itemize}

For each segment, we also retrieve the corresponding point cloud observation and action from the same timestep. We denote the observation-action pair for $\sigma_i$ as $\tau_i = (s_i, a_i)$ and use this representation for training the reward model.

We collect a dataset of labeled preferences $D = \{(\tau_o^k, \tau_1^k, y_k\}^N_{k=1}$, and discard all pairs where $y = -1$ before training. We train a reward function $r_\theta$ by minimizing the following loss:

\begin{equation}
\mathcal{L} = -\mathbb{E}_{(\tau_0, \tau_1, y) \sim \mathcal{D}} \left[
\mathbb{I}\{y = 0\} \log P_\theta[\tau_0 \succ \tau_1] +
\mathbb{I}\{y = 1\} \log P_\theta[\tau_1 \succ \tau_0]
\right]
\end{equation}

where $P_\theta[\tau_i \succ \tau_j]$ is the probability that $\tau_i$ is preferred over $\tau_j$, modeled using the Bradley-Terry formulation:

\begin{equation}
P_\theta[\tau_i \succ \tau_j] =
\frac{
\exp\left( r_\theta(\tau_i) \right)
}{
\exp\left( r_\theta(\tau_i) \right) + \exp\left( r_\theta(\tau_j) \right)
}
\end{equation}

The backbone of our reward model is a classification-type PointNet++ architecture, consisting of two set abstraction layers with radii of 0.05 and 0.1 and sampling ratio of 1.0 for both, followed by a global max pooling layer and a final MLP. The model is trained using the Adam optimizer, a learning rate of $1 \times 10^{-4}$, and a batch size of 64. Training is run for 1000 epochs or until convergence, whichever occurs first.

The total reward used to label the real-world dataset combines two components: a preference-based reward and a force-based penalty. The preference reward is provided by the learned reward model described above, while the force penalty discourages excessive contact force. The final reward is computed as a weighted sum of these two terms:

\begin{equation}
r_{\text{total}} = r_{\text{pref}} + w_{\text{force}} \cdot r_{\text{force}}
\end{equation}

where $r_{\text{pref}}$ is the output of the learned preference reward model, $r_{\text{force}}$ is a negative penalty based on contact force magnitude, and $w_{\text{force}}$ is a scalar set to be 0.1. $r_{\text{total}}$ is clamped to be between -1 and 1.

To discourage excessive contact force during dressing, we apply a penalty based on the magnitude of the applied force vector $\mathbf{f}$. The force magnitude is first normalized by dividing by 8~N, which corresponds to the 95th percentile of forces observed in the dataset, and then clipped to a maximum of 1. The final penalty is defined as:

\begin{equation}
r_{\text{force}} = -\min\left(1, \frac{\|\mathbf{f}\|}{8} \right)^2
\end{equation}

This formulation applies a normalized quadratic penalty on force magnitude, resulting in a smooth, increasing cost that discourages high-force interactions while still allowing gentle contact.

\section{Simulation Experiments}
\label{app:sim_setup}
\begin{figure}[h!] 
    \centering
    \includegraphics[width=0.7\linewidth]{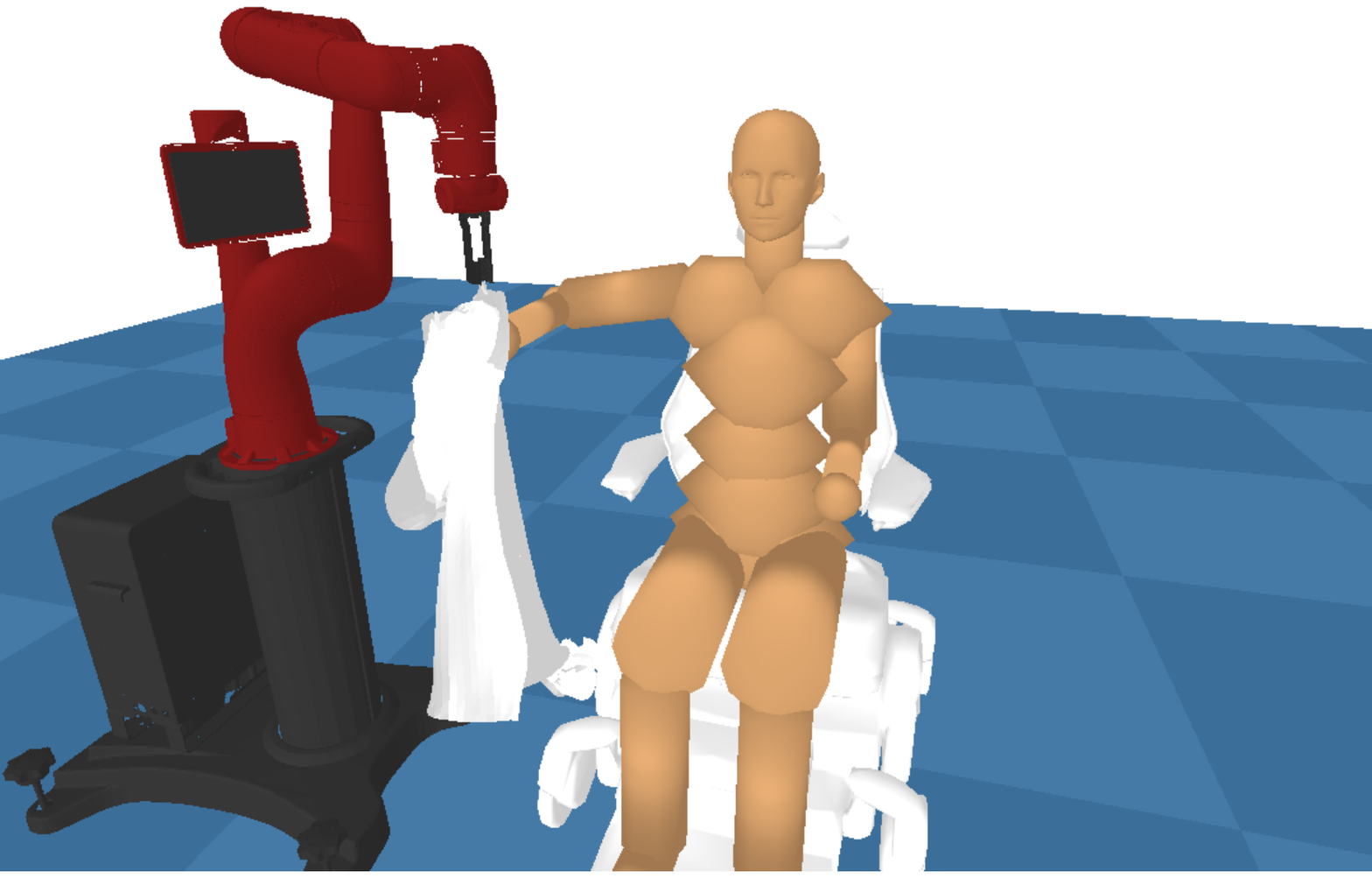}
    \caption{
     Simulation setup.
    }
    \label{fig:sim-setup}
\end{figure}

Figure~\ref{fig:sim-setup} illustrates our sim2sim transfer experiment setup in PyBullet. We use simplified cylindrical human meshes from Assistive Gym~\citep{erickson2020assistivegym} to approximate human bodies, as they are easily actuated and allow for consistent control. Our simulation includes four body sizes, three garments, and 14 arm motions. The 14 motions consist of seven base motions, each played both forward and in reverse. We set the maximum number of steps per trial to 250.

\begin{figure}[h!] 
    \centering
    \includegraphics[width=\linewidth]{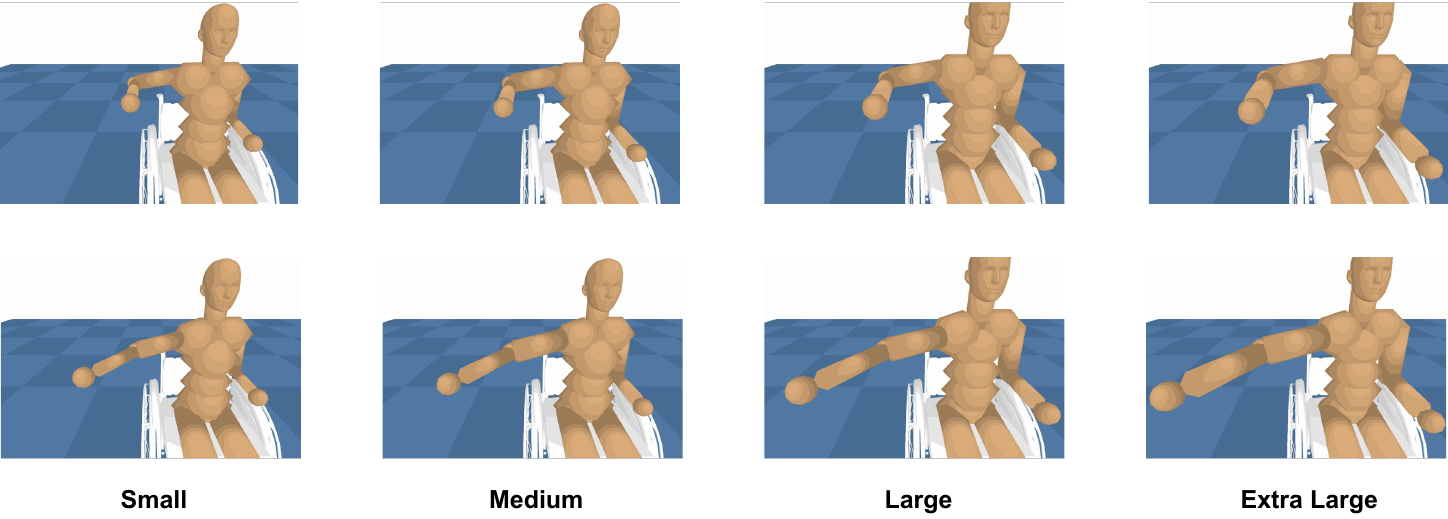}
    \caption{
     Body sizes in simulation.
    }
    \label{fig:sim-size}
\end{figure}

\textbf{Body sizes.} The small and medium body sizes are based on the default female mesh in Assistive Gym, while the large and extra large sizes are based on the default male mesh. Within each group (female or male), the only differences between the two sizes are the arm radius and length. We modify only the arm geometry—specifically, the length and radius of the upper arm and forearm—because the policy takes as input only the arm point cloud.
See Figure~\ref{fig:sim-size} for an illustration of the 4 different body sizes. 

Across the four body sizes, forearm radii range from 2.5 to 4.5 cm, forearm lengths from 20 to 28 cm, upper arm radii from 4 to 6 cm, and upper arm lengths from 24 to 30 cm.

\begin{figure}[h!] 
    \centering
    \includegraphics[width=\linewidth]{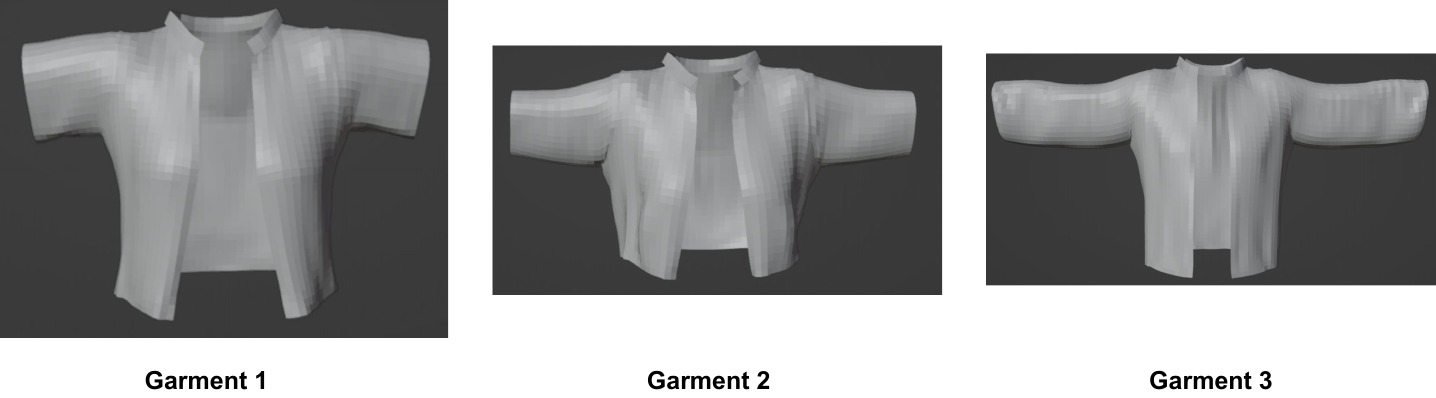}
    \caption{
     Dressing garments in simulation.
    }
    \label{fig:sim-garment}
\end{figure}

\textbf{Dressing garments.} As shown in Figure~\ref{fig:sim-garment}, we use three cardigans from the Cloth3D dataset~\citep{Bertiche2020cloth3d}, each with distinct geometries. The garments are scaled to realistic sizes appropriate for dressing.

\begin{figure}[h!] 
    \centering
    \includegraphics[width=\linewidth]{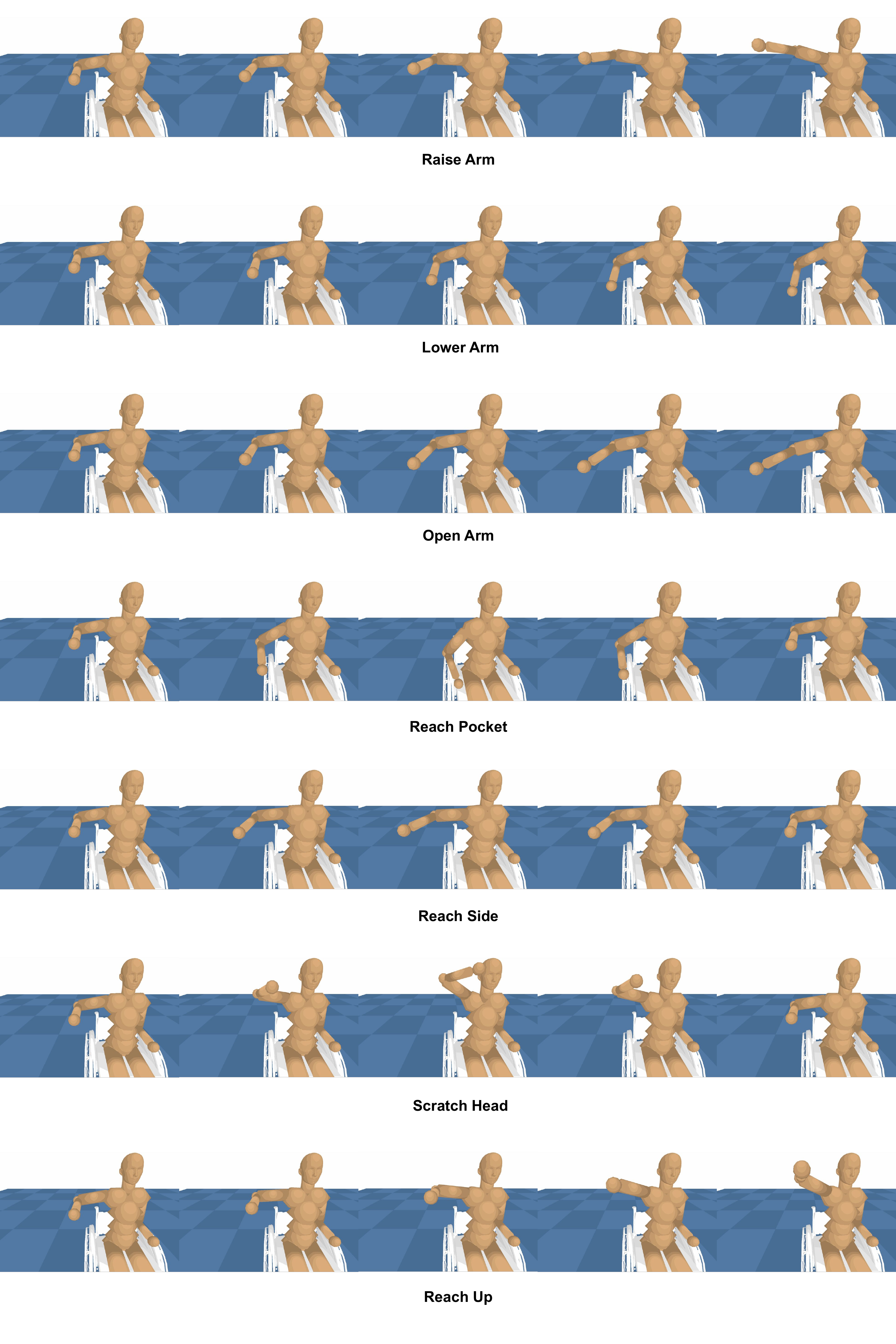}
    \caption{
     Base arm motions in simulation.
    }
    \label{fig:sim-motion}
\end{figure}

\textbf{Arm motions.} As shown in Figure~\ref{fig:sim-motion}, we define seven distinct arm motions and generate their reversed counterparts, resulting in 14 total motions. Each motion is defined by specifying a target arm pose using joint angles, and then performing linear interpolation from the initial position to the target to produce a complete trajectory. The first three motions—Raise Arm, Lower Arm, and Open Arm—and their reverses consist of 60 steps each. The remaining four motions—Reach Pocket, Reach Side, Scratch Head, and Reach Up—and their reverses each consist of 120 steps.

\subsection{Baseline Implementation}
\textbf{Vision-based Policy $\pi_{vis}$:} This baseline is trained in NVIDIA FleX using only visual observations. Details on the model architecture are described in Section~\ref{app:distill}.

\textbf{FCVP~\citep{sun2024force}:} We follow the implementation details outlined in~\citep{sun2024force}. This baseline uses the vision-based policy $\pi_{vis}$ to propose candidate actions, and then applies a force dynamics model trained in PyBullet to filter out actions that would exceed a predefined force threshold. The dynamic model takes as input: 1) the latent observation encoded by PointNet++, 2) the action vector, and 3) the force vectors from the previous five time steps. It outputs a prediction for the cumulative force that would be applied over the next five steps if the action were executed. We set the force threshold to be 40 N (i.e., 8 N per step). Actions with predicted cumulative force exceeding this threshold are discarded. From the remaining set of candidate actions, we select the one with the highest probability under $\pi_{\text{vis}}$. If all proposed actions exceed the threshold, we select the action with the lowest predicted cumulative force.

\textbf{Scratch-IQL (FiLM):} This baseline is trained from scratch using the dataset of 204 trials collected in PyBullet. It uses the same PointNet++ architecture and FiLM layers~\citep{perez2018film} to incorporate force information as in our method. The only difference is that the policy network in this baseline is not initialized from the vision-based policy pre-trained in NVIDIA FleX.

\textbf{Scratch-IQL (Concat):} This baseline is also trained from scratch using the dataset of 204 trials collected in PyBullet. It uses the same PointNet++ architecture as our method; however, instead of using FiLM conditioning, it concatenates the force magnitude directly to the robot end-effector point's position as an additional input feature. This adds an extra dimension to the feature vector, which originally only included one-hot indicators to distinguish between arm points, garment points, and robot end-effector points. As a result, this method is not compatible with our fine-tuning setup and is evaluated only when trained from scratch.

\textbf{Vision Fine-tuning:} This baseline only fine-tunes the vision network of $\pi_{vis}$ using the trajectories collected in PyBullet. It has no FiLM layers and does not incorporate force information.

\textbf{Force Fine-tuning:} This baseline follows the same approach and model architecture as our method, except the vision encoder of $\pi_{vis}$ is kept frozen.

\textbf{BC Fine-tuning:} This baseline follows the same model architecture as our method, but uses Behavioral Cloning as the underlying algorithm for fine-tuning with negative log likelihood loss.

\section{Real-World Experiments}
\subsection{Data Collection Study Procedure}
\textbf{Arm point cloud extraction.}
To extract the point cloud of only the participant’s right arm, we manually select a pixel on the depth image that corresponds to the shoulder at the beginning of the study. We then crop the point cloud to retain only points within a fixed range relative to the shoulder point: -45 to 5 cm in the x direction, -35 to 20 cm in the y direction, and -40 to 6 cm in the z direction. In future work, this manual step could be automated using a human pose estimator.

\textbf{Dressing trial length.} We set the maximum number of steps allowed per trial to 80, with each trial typically lasting between one and two minutes. A full study session generally takes 1 to 1.5 hours, including time for showing participants demonstration videos, changing garments, and providing rest breaks.

\textbf{Scripts for participant.} We read and show the following script to each participant at the beginning of each study session to ensure familiarity with the study procedure:

\textit{Thank you for participating in our study to evaluate a robot-dressing system! The robot will dress the garment on your right arm. Here is a quick overview of what to expect during the study: You will first read and sign a consent form, and then fill out a demographic questionnaire. Before we start, we will take some measurements of your arm, including forearm length, upper arm length, and the arm circumference. We will then start the study. There will be 24 dressing trials using three garments and eight arm motions. Each trial will feature a unique combination of these elements. For each trial, you will be asked to perform simple arm motions, such as moving your arm up and down. We will show you a demo video of the motion, and if needed, we can demonstrate the motion for you to ensure clarity. Please perform the motion slower than you naturally would. Once we indicate it’s time to start, you will perform the arm motion while the robot dresses you. Very occasionally, the robot’s gripper might make contact with you during the dressing process. If at any point you feel uncomfortable, please let us know, and we can stop the trial. Occasionally there might be operational issues during a trial. If that happens, we will repeat those trials as needed. After each trial, please keep your arm still while we take some measurements to evaluate the dressing performance. After that, you can rest your arm and fill out a questionnaire about your experience. Feel free to let us know if you need a break at any time. After every 8 trials with a garment, we will change the garment, and you will have the chance to rest. Thank you for your cooperation and participation. We appreciate your help in this study.}

\subsection{Evaluation Study Procedure}
The evaluation study follows a similar procedure to the data collection study, with the main differences being the number of dressing trials and the garments and arm motions used. In the evaluation, we use two new garments that are not part of the data collection, along with seven arm motions, including three that are not used during data collection.

During the data collection study, we use garments with a variety of geometries, such as wide sleeves and elastic fabrics that make dressing easier. The arm motions used in data collection cover a wide range, but some are random or artificial (e.g., bending the arm) rather than natural, purposeful motions that people might perform during everyday dressing (e.g., rubbing the face).

To evaluate our method on more realistic and challenging scenarios, we purchase two new garments with long, narrow, and non-elastic sleeves from a nearby shopping center. We also replace some of the training motions with meaningful, natural actions--such as taking a phone out of a pocket and scrolling on the screen, and waving--that are more likely to occur in real-world settings. 

For each participant, we run 11 trials per garment, totaling 22 trials. Of the 11 trials, seven use our method (covering all seven motions), and the remaining four are split between two baselines, with each baseline evaluated on two randomly selected motions. The motion assignments for baseline trials are counterbalanced across participants to ensure that, by the end of all the study sessions, each motion is used approximately the same number of times for each baseline.

\subsection{Evaluation Study Analysis}
\label{app:more-eval-analysis}
\begin{figure}[h!]
\centering
\begin{minipage}[c]{0.4\linewidth}
    \centering
    \includegraphics[width=\linewidth]{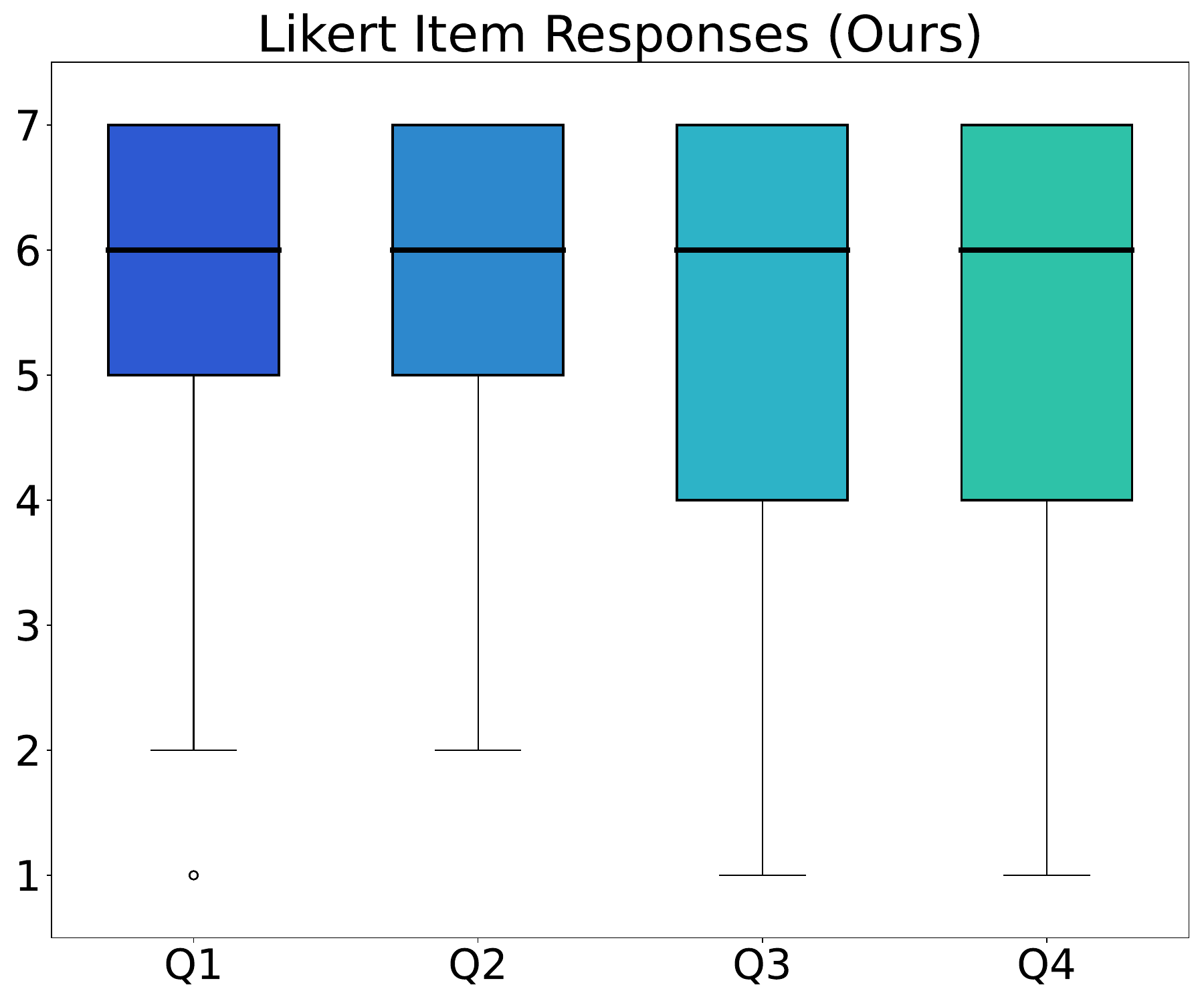}
\end{minipage}%
\begin{minipage}[c]{0.6\linewidth}
    \centering
    \begin{tabular}{l     
        >{\centering\arraybackslash}m{0.25\linewidth}
        >{\centering\arraybackslash}m{0.25\linewidth}
     }

        \toprule
         & \shortstack{Upper Arm\\Dressed Ratio} & \shortstack{Whole Arm\\Dressed Ratio}\\
        \midrule
        Lower Arm  & 0.40 & 0.60\\
        Rub Nose  & 0.74 & 0.84\\
        Wave  & 0.87 & 0.93\\
        Use Phone  & 0.86 & 0.92\\
        Receive Bottle  & 0.84 & 0.92\\
        Scratch Head  & 0.73 & 0.86\\
        Improvise  & 0.74 & 0.86\\
        \bottomrule
    \end{tabular}
\end{minipage}
\caption{Likert item responses (left) and average arm dressed ratios (right) for our method, evaluated on all 168 trials (not just the subset shown in Figure 4 in the main paper).} 
\label{fig:human_study_ours}
\end{figure}
A Friedman test is conducted to examine whether participants' ratings differed across the three methods for each of the four Likert-scale questions. For all four questions, the results were statistically significant (p $<$ 0.05), indicating that participants' perceptions varied significantly depending on the method used. To further explore these differences, we conduct Wilcoxon signed-rank tests for pairwise comparisons between our method and each baseline. We find significant differences in all comparisons across the four questions, except when comparing our method with the vision-based method for Q4, where the difference is not statistically significant.

We now analyze the performance of our method across the 168 trials conducted with 12 participants. Figure~\ref{fig:human_study_ours} shows the upper arm dressed ratio of our method across all arm motions. ``Improvise" refers to the condition where participants were allowed to move their arm freely. Notably, our method achieves relatively consistent performance across most arm motions but shows a significant drop for the ``Lower Arm" motion. One possible explanation is that ``Lower Arm" results in severe occlusion from the camera view: as participants lower their arm, it becomes almost completely covered by the garment, leading to limited visual information for the network. Additionally, some participants find it difficult to lower their arm while being dressed in a long-sleeved garment and often apply large forces to pull both the garment and the robot end-effector downward. Such high-force interactions may not be well represented in the dataset used for training, as the garments used during data collection have wide sleeves or elastic textures. This mismatch could lead to out-of-distribution robot behaviors during execution.

\subsection{Evaluation Study Failure Cases}
\begin{figure}[h!] 
    \centering
    \includegraphics[width=\linewidth]{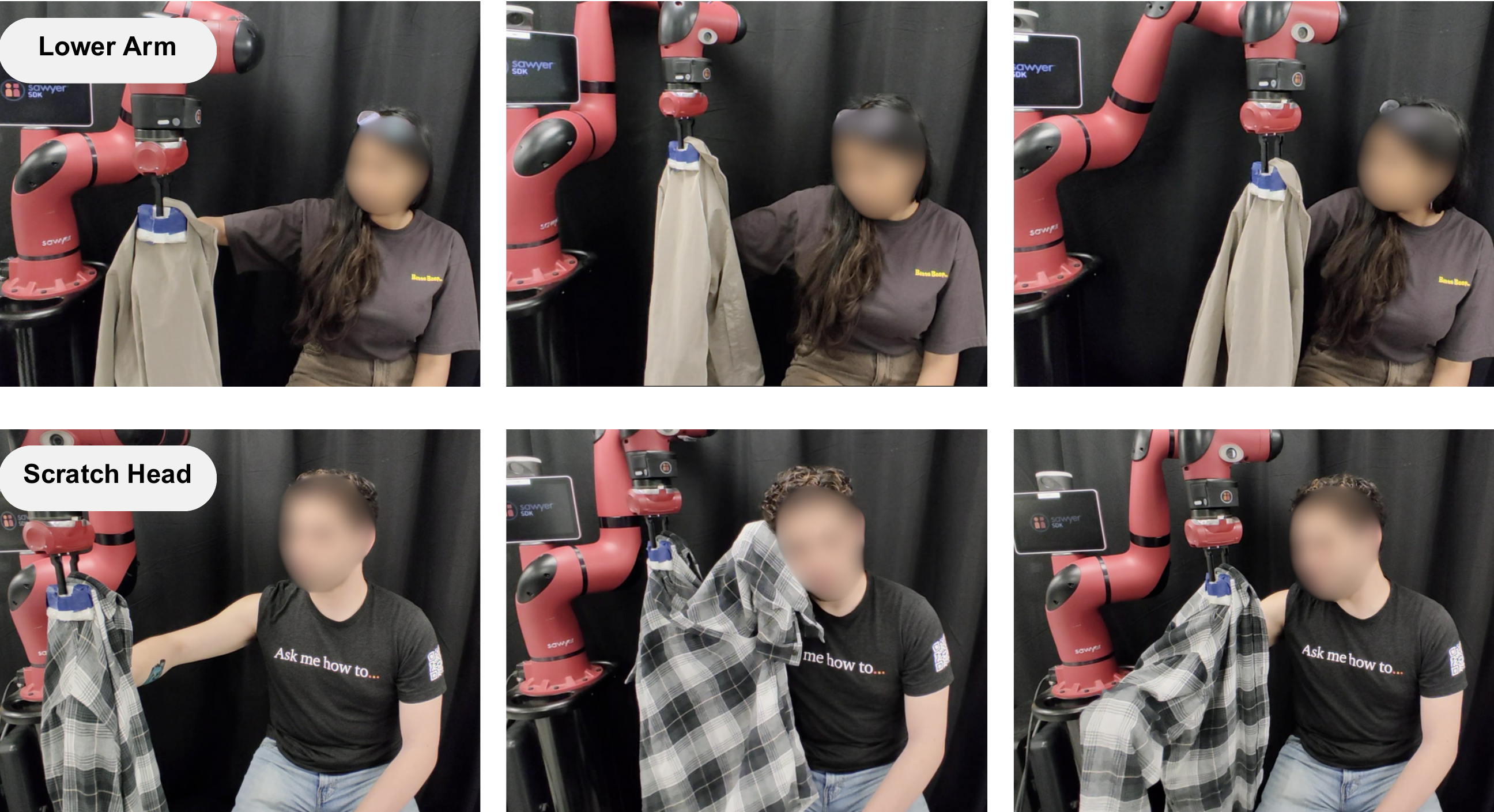}
    \caption{
     Failure cases of our system in the human study. (Top) garment gets caught on the elbow as the participant performs the ``Lower Arm" motion. (Bottom) the policy actions turn inward too early and stop making progress towards the upper arm.
    }
    \label{fig:failure}
\end{figure}

Figure~\ref{fig:failure} shows two failure cases of our method. In the first case, the policy fails to adapt to the participant’s arm-lowering motion, causing the garment to get caught beneath the elbow. As discussed in Section~\ref{app:more-eval-analysis}, this may be due to a combination of severe occlusion from the camera view and out-of-distribution high-force interactions. 

In the second case, the policy initiates the turning motion while the garment is still on the forearm, instead of waiting until it reaches the elbow. As a result, the policy stops making dressing progress, moving horizontally in front of the participant. From the camera view, it is difficult to visually localize the elbow, as the scratch head motion causes even greater occlusion due to the garment being stretched. By the time the participant returns to the initial arm position, the gripper has already moved in front of the body, requiring a significant recovery to return to the correct trajectory.

\begin{figure}[h!] 
    \centering
    \includegraphics[width=0.85\linewidth]{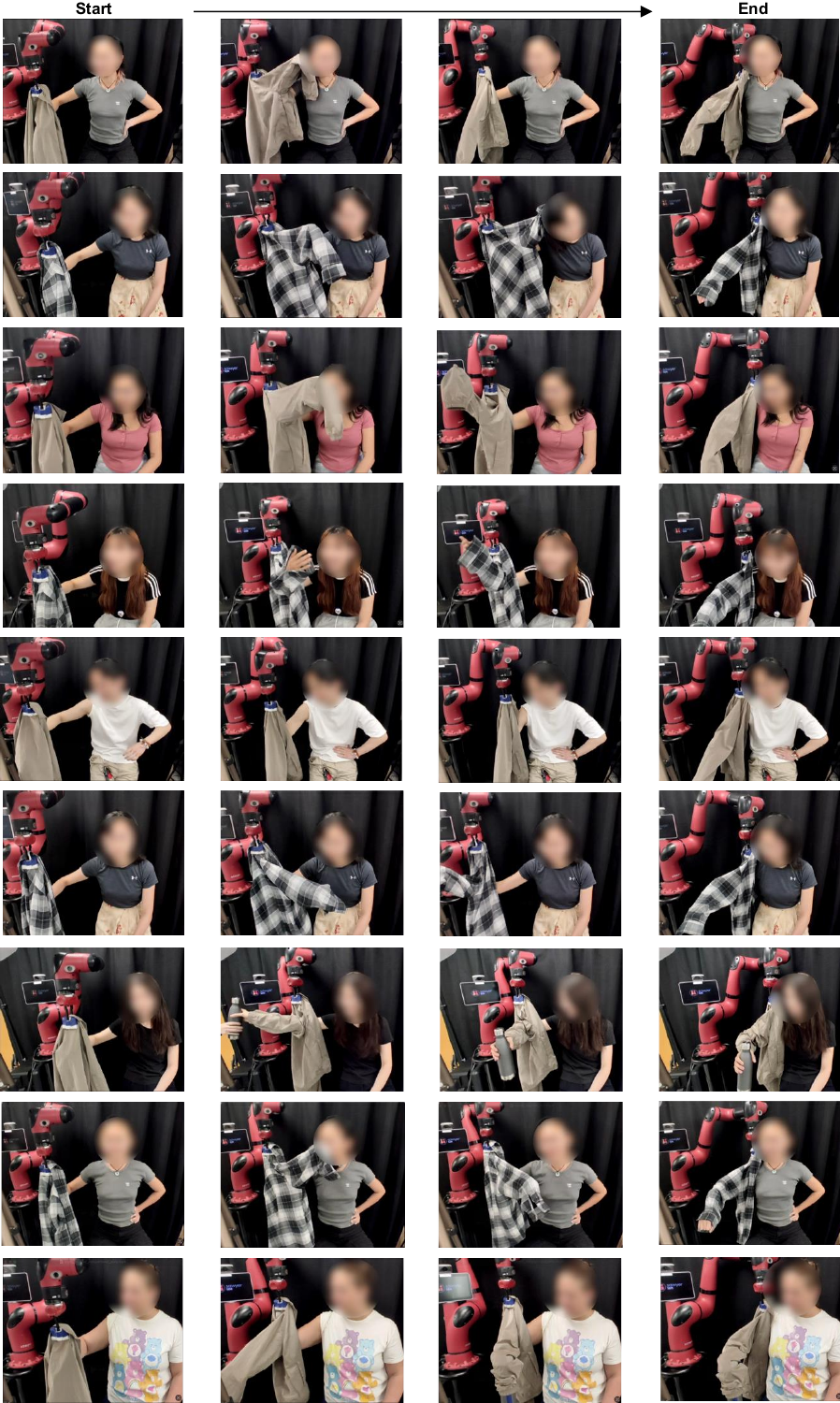}
    \caption{
     Additional successful dressing trials using our method, not shown in the main paper.
    }
    \label{fig:more-figures}
\end{figure}

\end{document}